\documentclass[journal]{IEEEtran}
\usepackage{caption}
\usepackage{tikz}
\usepackage{amssymb}
\usepackage{amsthm}
\theoremstyle{plain}

\usepackage{booktabs}
\usepackage{amsmath}
\usepackage{algpseudocode}
\usepackage{caption}
\usepackage{tabulary}
     
\usepackage{enumitem}
\usepackage{booktabs}

\usepackage{graphicx}
\usepackage{subcaption}
\usepackage[export]{adjustbox}

\usepackage{adjustbox}
\usetikzlibrary{positioning, shapes.geometric}
\usepackage{xcolor}

\usepackage{mathtools}

\usepackage{array,booktabs,tabularx}
\usepackage{makecell}
\newcolumntype{P}[1]{>{\centering\arraybackslash}p{#1}}

\usepackage{cite}

\usepackage{graphicx}
\usepackage{amssymb}
\usepackage{floatrow}
\usepackage{amsmath}
\usepackage{lipsum}
\usepackage{multicol}
\usepackage{tikz}
\newcommand{\nonl}{\renewcommand{\nl}{\let\nl\oldnl}}

\usetikzlibrary{arrows,shapes,positioning,shadows,trees}

\usepackage{algorithm}

\tikzset{
  basic/.style  = {draw, text width=2cm, drop shadow, font=\scriptsize, rectangle},
  root/.style   = {basic, rounded corners=2pt, thin, align=center,
                   fill=white!30},
  level 2/.style = {basic, rounded corners=4pt, thin,align=center, fill=white!60,
                   text width=5em},
  level 3/.style = {basic, thin, align=left, fill=white!60, text width=5em}
}

\newcommand{\CASE}[1]{\STATE \textbf{case} #1\textbf{:} \begin{ALC@g}}
\newcommand{\ENDCASE}{\end{ALC@g}}

\newcommand{\DEFAULT}{\STATE \textbf{default:} \begin{ALC@g}}
\newcommand{\ENDDEFAULT}{\end{ALC@g}}
\newcommand{\DEFAULTLINE}[1]{\STATE \textbf{default:} }

\makeatletter
\g@addto@macro\normalsize{%
  \setlength\abovedisplayskip{6pt plus 2pt minus 2pt}%
  \setlength\belowdisplayskip{6pt plus 2pt minus 2pt}%
  \setlength\abovedisplayshortskip{4pt plus 2pt minus 2pt}%
  \setlength\belowdisplayshortskip{4pt plus 2pt minus 2pt}%
}
\makeatother
\usepackage[font=small,skip=4pt]{caption}

\usepackage{titlesec}
\titlespacing*{\section}{0pt}{8pt}{4pt}
\titlespacing*{\subsection}{0pt}{6pt}{2pt}

\ifCLASSINFOpdf
\else
\fi
\begin{document}
%
\title{CLAIRE: Compressed Latent Autoencoder for Industrial Representation and Evaluation — A Deep Learning Framework for Smart Manufacturing}
%
%

\author{Mohammadhossein Ghahramani,~\IEEEmembership{Senior Member,~IEEE,} and Mengchu Zhou,~\IEEEmembership{Fellow,~IEEE}

\thanks{}
\thanks{}
\thanks{M. Ghahramani is with Birmingham City University, UK, (e-mail: mohammadhossein.ghahramani@bcu.ac.uk)}

\thanks{M. C. Zhou is with the Helen and John C. Hartmann Department of Electrical and Computer Engineering, New Jersey Institute of Technology, Newark, NJ 07102, USA (e-mail: zhou@njit.edu).}

}

%
%

\markboth{}%
{Shell \MakeLowercase{\textit{et al.}}: Bare Demo of IEEEtran.cls for IEEE Journals}
%



\maketitle

\begin{abstract}
Accurate fault detection in high-dimensional industrial environments remains a major challenge due to the inherent complexity, noise, and redundancy in sensor data. This paper introduces CLAIRE, i.e., a hybrid end-to-end learning framework that integrates unsupervised deep representation learning with supervised classification for intelligent quality control in smart manufacturing systems. It employs an optimized deep autoencoder to transform raw input into a compact latent space, effectively capturing the intrinsic data structure while suppressing irrelevant or noisy features. The learned representations are then fed into a downstream classifier to perform binary fault prediction. Experimental results on a high-dimensional dataset demonstrate that CLAIRE significantly outperforms conventional classifiers trained directly on raw features.  Moreover, the framework incorporates a post hoc phase, using a game-theory-based interpretability technique, to analyze the latent space and identify the most informative input features contributing to fault predictions. 
The proposed framework highlights the potential of integrating explainable AI with feature-aware regularization for robust fault detection. The modular and interpretable nature of the proposed framework makes it highly adaptable, offering promising applications in other domains characterized by complex, high-dimensional data, such as healthcare, finance, and environmental monitoring.

\makeatletter{\renewcommand*{\@makefnmark}{}
\footnotetext{}\makeatother}

\end{abstract}


%
\IEEEpeerreviewmaketitle

\section{Introduction}\label{section.intro}
%
%
%
%

\IEEEPARstart{M}{aintaining} product integrity and promptly identifying faults are essential objectives in contemporary industrial systems. The integration of IoT-based sensors and advanced monitoring infrastructure has resulted in the rapid accumulation of high-dimensional process data. Although these data streams hold promise for detecting anomalies, their high noise levels and redundant features reduce the efficacy of conventional machine learning models when applied to unprocessed inputs. As manufacturing operations grow in complexity, there is an increasing demand for robust and adaptive data-driven methods to support quality assurance. Traditional supervised learning models often underperform in these settings due to their limitations in managing the dimensionality and interdependencies present in sensor data.

Recent progress in computational power, along with the emergence of innovative learning paradigms, has brought substantial improvements to many industrial applications. These advancements have paved the way for more effective strategies to address complex problems. Among the most influential developments is the rise of deep learning, which has reshaped the way high-dimensional data is handled. Deep neural networks have proven highly effective in domains such as image analysis, speech recognition, and industrial system monitoring. These methods have also gained prominence in feature extraction due to their ability to automatically learn patterns from raw data. 

Convolutional Neural Networks (CNNs) and Recurrent Neural Networks (RNNs) are widely used in image and sequence modeling, respectively \cite{Huang2025,Tang2024,LeiZhou2024}. CNNs excel at capturing spatial patterns in structured data like images, while RNNs are effective for sequential inputs and temporal dependencies. However, their reliance on spatial or temporal structure limits their applicability in smart manufacturing, where sensor data is often high-dimensional and unstructured. In such settings, autoencoders provide a more suitable alternative. Designed for unsupervised learning and dimensionality reduction, they can effectively filter noise and redundancy, uncovering latent features that capture the essential structure of complex industrial data. This property of autoencoders makes them particularly useful for fault detection in smart manufacturing. By learning compact and informative representations, they can enable us to identify subtle patterns linked to equipment faults or product defects.

This study addresses the challenge of accurate product quality prediction in smart manufacturing environments characterized by high-dimensionality and redundant features. When traditional classifiers are applied directly to such raw inputs, their performance is often compromised due to the curse of dimensionality and the presence of non-informative features. To overcome these limitations, we propose CLAIRE (i.e., Compressed Latent Autoencoder for Industrial Representation and Evaluation), as a hybrid end-to-end framework that integrates unsupervised deep representation learning with supervised classification. Specifically, an optimized deep autoencoder is implemented to learn compact and informative latent representations that capture the intrinsic structure of the input data while filtering out redundancy. These learned features are subsequently used as inputs to a downstream classifier for binary fault prediction. By decoupling the feature extraction and classification phases, the proposed approach enhances generalization performance and improves predictive robustness in complex industrial settings. To interpret the latent space including learned features, we employ an advanced interpretability technique, i.e., a game theory-based approach, to gain insights into which features are most influential in the model’s decision-making process. The name "CLAIRE" reflects our proposed framework’s core philosophy of clarity and transparency in AI. Notably, while deep learning models are often perceived as black boxes, CLAIRE is designed with transparency in mind--offering not just predictive performance, but also a clearer understanding of the learned representations. This work intends to make the following new contributions:
\begin{enumerate}
    \item It proposes CLAIRE, an end-to-end hybrid framework that couples unsupervised deep representation learning with supervised fault classification. Unlike existing approaches that treat feature extraction and classification as disjoint steps, CLAIRE integrates an optimized denoising autoencoder with a kernel-based SVM, yielding compact latent embeddings that are directly optimized for discriminative fault detection in high-dimensional manufacturing data;
    
    \item It introduces a novel joint optimization strategy that adaptively tunes both reconstruction and classification objectives through dynamic learning rate scheduling. This includes the use of latent variance regularization, which promotes well-separated and informative embeddings; and
    
    \item It develops an explainability phase that applies game-theoretic analysis to latent space. Beyond simple feature attribution, this phase reveals i) how the learned embeddings achieve clear class separation, and ii) which raw input features most strongly drive this separation. This provides actionable and domain-relevant insights that improve model transparency and increase trust in its deployment for fault diagnosis in industrial systems.
\end{enumerate}

The remainder of this paper is structured as follows. Section \ref{relatedWork} reviews existing literature on fault detection and deep learning techniques. The proposed hybrid model and its architectural components are detailed in Section \ref{section.Model}. Experimental setup, evaluation metrics, and performance results are presented in Section \ref{section.Results}. Finally, Section \ref{section.conclusion} concludes the paper and outlines potential directions for future research.

\section{Related Work}\label{relatedWork}
Traditional Statistical Process Control (SPC) methods, while effective for low-dimensional monitoring, often fall short in handling the complexity and scale of modern sensor-based manufacturing data. Because of this, recent years have witnessed a growing interest in leveraging Machine Learning (ML) and Artificial Intelligence (AI) techniques for industrial fault detection and quality control. 

However, the sheer volume and dimensionality of sensor data in these settings introduce new challenges, such as redundancy, noise, and overfitting. To address these issues, feature selection has become essential for enhancing the performance and interpretability of fault detection models. By selecting the most relevant features from high-dimensional sensor data, models become more robust to noise and computationally efficient. This section categorizes feature selection methods into traditional and advanced approaches, highlighting their evolution and limitations.

\subsection{Traditional Approaches}
Conventional feature selection techniques are broadly categorized into filter, wrapper, and embedded methods \cite{Tanzhou2022, Liuzhou2020}. Filter methods, such as Mutual Information Maximization (MIM), Minimum Redundancy Maximum Relevance (MRMR), and Fast Correlation-Based Filter (FCBF), assess features based on statistical metrics like correlation or mutual information \cite{Hochma2024}. These methods are computationally efficient and model-agnostic but typically evaluate each feature independently, overlooking interactions and non-linear relationships among features \cite{Mandal2024}. Wrapper methods, such as Recursive Feature Elimination (RFE), involve iterative training of a predictive model to assess subsets of features based on classification performance. While they often yield higher prediction accuracy, they are computationally intensive, especially in high-dimensional low-sample-size (HDLSS) settings, and are prone to overfitting. Embedded methods, including LASSO regression and decision tree-based models, incorporate feature selection during model training by introducing regularization or through hierarchical feature evaluation. Although more scalable than wrapper approaches, embedded methods may still fall short when modeling complex, non-linear dependencies and often depend on strong assumptions about the underlying data structure \cite{Lee2025}. Collectively, these traditional approaches struggle with scalability, interaction modeling, robustness to noise, and generalizability in modern industrial applications involving high-dimensional, noisy, and heterogeneous sensor data \cite{Lin2024, Li2023}.

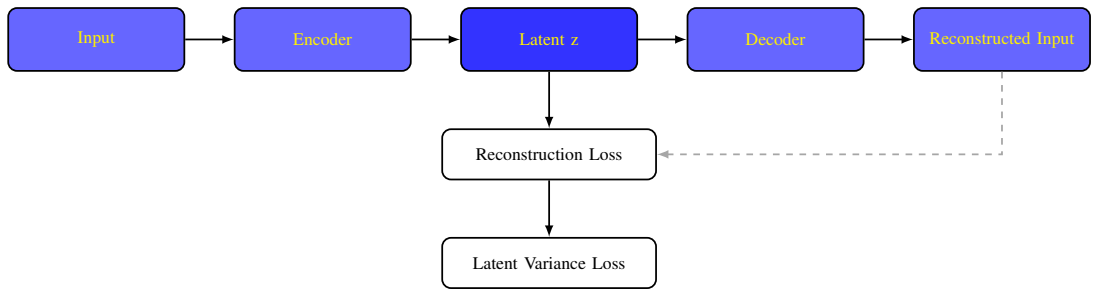
\begin{figure*}[htbp]
    \centering
    \resizebox{0.8\linewidth}{!}{%
    \begin{tikzpicture}[
        >=latex,
        node distance=3.6cm,
        font=\footnotesize,
        every node/.style={align=center},
        block/.style={rectangle, draw, rounded corners, minimum height=10mm, minimum width=28mm, thick, fill=blue!60, text=yellow},
        latent/.style={rectangle, draw, rounded corners, minimum height=10mm, minimum width=28mm, thick, fill=blue!80, text=yellow},
        loss/.style={rectangle, draw, rounded corners, minimum height=8mm, minimum width=34mm, thick, fill=white},
        arrow/.style={->, thick},
        dashedarrow/.style={->, thick, dashed, gray!70}
    ]

    \node[block] (input) {Input};
    \node[block, right of=input] (encoder) {Encoder};
    \node[latent, right of=encoder] (latent) {Latent z};
    \node[block, right of=latent] (decoder) {Decoder};
    \node[block, right of=decoder] (reconstructed) {Reconstructed Input};

    \node[loss, below=0.9cm of latent] (recon_loss) {Reconstruction Loss};
    \node[loss, below=0.9cm of recon_loss] (latent_loss) {Latent Variance Loss};

    \draw[arrow] (input) -- (encoder);
    \draw[arrow] (encoder) -- (latent);
    \draw[arrow] (latent) -- (decoder);
    \draw[arrow] (decoder) -- (reconstructed);

    \draw[arrow] (latent) -- (recon_loss);
    \draw[arrow] (recon_loss) -- (latent_loss);
    \draw[dashedarrow] (reconstructed.south) |- (recon_loss.east);

    \end{tikzpicture}
    }
    \caption{Encoder--Decoder Structure with Latent Space Regularization. This architecture incorporates reconstruction loss and latent variance loss. Note that, a supervised learning layer, optimized using cross-entropy loss and entropy regularization, and a latent exploration module (both not shown here) follow the encoder. These components are discussed in detail throughout the paper and illustrated in Fig.~2.}
    \label{fig:autoencoder_structure}
\end{figure*}

\subsection{Advanced Approaches}
To address the limitations of traditional feature selection techniques, recent studies have increasingly explored metaheuristic optimization and deep learning-based approaches for industrial data modeling. Metaheuristic methods, such as Evolutionary Algorithms (EAs), leverage population-based global search strategies to identify informative feature subsets in high-dimensional spaces \cite{Ghahramani2021,Chen2025Workload,Deng2025Evolutionary,Ghahramani2025Gentelligent}. To alleviate their high computational cost, surrogate-assisted variants have been introduced, where fitness evaluations are approximated using probabilistic models such as Gaussian processes \cite{Liu2022,Ahadzadeh2023}. Reinforcement Learning (RL) has also been investigated as an adaptive alternative, framing feature selection as a sequential decision-making process. For example, Fan et al.~\cite{Fan2023} propose an interactive RL-based feature selection strategy guided by classifier feedback, although scalability and training complexity remain key challenges in industrial settings.

Alongside these optimization-based paradigms, deep learning has emerged as a powerful tool for automated feature extraction and representation learning. Deep neural networks have demonstrated remarkable success in learning hierarchical and non-linear feature representations across a wide range of domains, including image analysis, speech recognition, and industrial system monitoring. In manufacturing and process industries, deep learning models are particularly attractive due to their ability to operate directly on raw or minimally processed sensor data and to capture complex dependencies among process variables.

Among deep learning techniques, autoencoder-based models have gained significant attention for unsupervised representation learning in industrial environments. Autoencoders map high-dimensional inputs into compact latent representations while preserving essential information, making them well suited for noisy, redundant, and highly correlated sensor data commonly encountered in smart manufacturing systems \cite{Yasutomi2025,Wang2025}. Variants such as denoising autoencoders (DAEs) further enhance robustness by explicitly learning to reconstruct clean signals from corrupted inputs \cite{Tu2022}. More advanced architectures have been proposed to capture temporal or structural dependencies, including self-attentive convolutional autoencoders for wind turbine monitoring \cite{Yan2022}, position-encoding denoising autoencoders for missing-data imputation, and deep residual networks for data-driven quality prediction in industrial processes \cite{Xiao2025Fault}. Autoencoder-based frameworks have also been successfully applied to other industrial and infrastructure-oriented tasks. For instance, Meng et al. \cite{Meng2023HVAE} propose a hierarchical variational autoencoder (VAE) for physical-layer authentication in Industrial Internet of Things (IIoT) environments, demonstrating how hybrid AE--VAE architectures can learn robust representations under limited and uncertain training data. Similarly, Shahid et al. \cite{Shahid2024VAE} investigate VAEs and recurrent autoencoder models for anomaly detection and knowledge transfer in electricity and district heating consumption, highlighting the effectiveness of reconstruction-based learning for generalizable industrial anomaly detection. These studies collectively illustrate the versatility of autoencoder-based frameworks in handling diverse industrial data challenges. Beyond reconstruction-focused learning, recent research has also explored probabilistic and information-theoretic extensions of autoencoders, such as information bottleneck formulations \cite{Cho2024Anomaly} and representation disentanglement techniques \cite{Chang2025Causal,Li2024ODCL}. While these approaches are effective for generative modeling or factorized representation learning, their objectives are not explicitly aligned with discriminative fault detection or latent-space compactness under industrial noise conditions. In parallel, self-supervised and contrastive learning frameworks \cite{Cho2024Supervised,Hu2025Self} have been proposed to learn general-purpose representations through instance-level discrimination; however, such methods typically rely on carefully designed data augmentations or structural assumptions that are not naturally available in tabular industrial sensor datasets.

In this work, we build upon the strengths of encoder--decoder-based representation learning while explicitly addressing the requirements of industrial fault detection. We propose an optimized denoising autoencoder architecture that integrates reconstruction loss with a latent variance regularization term, aiming to shape the geometry of latent space toward compact and stable embeddings. This unsupervised representation learning phase forms the foundation of the framework and is complemented by a downstream kernel-based classifier and a latent exploration layer for interpretability. By combining discriminative latent regularization with a game-theoretic explanation mechanism, the proposed framework seeks to balance predictive performance, robustness, and transparency. To clarify the relationship between CLAIRE and some representative deep representation learning paradigms, a conceptual comparison is summarized in Table~\ref{tab:comparison}.

\begin{table*}[t]
\centering
\caption{Conceptual comparison of representative deep representation learning frameworks for industrial fault detection.}
\label{tab:comparison}
\footnotesize
\begin{tabular}{@{}p{4cm} p{13.5cm}@{}}
\toprule
\textbf{Method Category} & \textbf{Objective and Key Characteristics} \\
\midrule

Standard Autoencoder (AE) &
Learn compact representations by minimizing reconstruction error; denoising AEs improve robustness but latent space geometry is not explicitly constrained for discriminative separability or compactness.\\
\addlinespace[4pt]

Variational AE &
Introduce probabilistic latent variables with KL-based regularization to enforce prior matching and promote disentanglement; primarily designed for generative modeling rather than fault-specific discrimination. \\
\addlinespace[4pt]

Information Bottleneck AE &
Optimize a trade-off between reconstruction and information compression; focus on minimal sufficient representations rather than explicit latent compactness or class separation.\\
\addlinespace[4pt]

Self-supervised and Contrastive Learning &
Learn representations via instance-level contrastive objectives and data augmentations; often require structured inputs or domain-specific augmentation strategies, limiting applicability to tabular industrial sensor data. \\
\addlinespace[4pt]

\textbf{CLAIRE (Proposed)} &
Jointly optimizes denoising reconstruction and latent variance regularization to explicitly shape compact and stable latent embeddings for fault discrimination, complemented by kernel-based classification and latent-level interpretability.\\
\bottomrule
\end{tabular}
\end{table*}

\section{Problem Formulation and Proposed Approach}\label{section.Model}
\subsection{Dataset and Problem Overview}
This work utilizes two publicly available benchmark datasets commonly used in the domain of manufacturing process monitoring, i.e., SECOM and Tennessee Eastman Process (TEP). SECOM contains operational data collected during a real semiconductor manufacturing process with a target variable indicating the production outcome, i.e., success or failure. It includes more than 500 features corresponding to various sensor readings. The target variable is binary, indicating whether the manufacturing process results in success (1) or failure (0). Some features contain null values, making imputation impractical; thus, these features are excluded from the dataset. TEP is a widely used simulated chemical process dataset that represents a realistic industrial process with multiple operating modes and both normal and faulty conditions \cite{Lomov2021}. It contains 52 process variables recorded as multivariate data, along with labeled fault classes, which makes it suitable for evaluating fault detection and diagnosis methods. Both datasets exhibit substantial class imbalance, which is a well-documented challenge in industrial fault detection. In SECOM, the proportion of defective samples is smaller than that of successful production runs, and in TEP, certain fault classes occur less frequently than normal operating conditions. To mitigate the bias induced by this imbalance, we employ stratified sampling, ensuring that minority classes are adequately represented during model training. These measures improve the reliability of the evaluation and prevent the model from being biased toward the majority class.

Denote the dataset as matrix \( X \in \mathbb{R}^{n \times d} \), where each row corresponds to an individual observation and each column to a specific feature, such as a sensor reading or process variable. Here, \( n \) denotes the number of observations, and \( d \) the number of measured features. This matrix serves as the input to the predictive models used in this study and is structured as follows:

\[
X = \begin{bmatrix}
x_{11} & x_{12} & \cdots & x_{1d} \\
x_{21} & x_{22} & \cdots & x_{2d} \\
\vdots & \vdots & \ddots & \vdots \\
x_{n1} & x_{n2} & \cdots & x_{nd}
\end{bmatrix}
\]

where \( x_{ij} \) is the value of the \( j \)-th feature in the \( i \)-th observation. Alongside this, the target vector \( Y \in \mathbb{R}^{n} \) holds the corresponding class labels for each observation, where \( Y = [y_1, y_2, \dots, y_n]^T \), and each label \( y_i \in \{0, 1\} \), where 1 denotes a "Success" and 0 indicates a "Failure."

Prior to model training, several preprocessing steps are carried out on the dataset. These include: 1) Handling missing data; 2) Addressing class imbalance using techniques such as oversampling \cite{YanZhao2024}; 3) Outlier detection to ensure the quality and reliability of the data.

Matrix \( X \) serves as the input of the proposed model. The input data is fed into the encoder. The encoder learns to map high-dimensional input data into lower-dimensional latent space, where the most informative features are captured. During training, the decoding component ensures that the reconstruction loss is minimized. To ensure the quality of the learned representations, we employ Reconstruction Loss and Latent Variance Loss to minimize the reconstruction error and penalize excessive variation within each class's latent space, respectively. Additionally, Dropout Regularization (DR) and Batch Normalization (BN) are applied to regularize the model and improve its generalization capabilities. DR randomly deactivates neurons during training, which prevents co-adaptation of latent features and reduces overfitting in high-dimensional settings. BN stabilizes and accelerates training by normalizing intermediate activations, thereby mitigating internal covariate shift and improving robustness to noisy sensor data. Together, these techniques are particularly important in industrial fault detection, where datasets are often noisy and prone to overfitting due to high dimensionality. The features extracted from \( X \) in the latent space are then used to train classifiers, where the goal is to optimize the prediction accuracy based on the input features. The data from \( X \) is split into training and test sets to validate the effectiveness of the models in predicting unseen data. The analysis and evaluation are carried out by using different performance metrics. To further evaluate and interpret the extracted features, we integrate a game-theory-based interpretability technique. This enables us to analyze the latent space and identify the most influential features in a fault prediction process.

\begin{figure*}[t]
\centering
\resizebox{\textwidth}{!}{%
\begin{tikzpicture}[
    node distance=1.7cm and 1.8cm,
    every node/.style={font=\Large},
    layer/.style={
        rectangle, draw,
        minimum width=4.5cm,
        minimum height=2.2cm,
        rounded corners,
        align=center,
        font=\Large
    },
    latent/.style={
        rectangle, draw, thick,
        minimum width=4.5cm,
        minimum height=2.2cm,
        rounded corners,
        align=center,
        font=\Large
    },
    classifier/.style={
        rectangle, draw, dashed, thick,
        minimum width=4.5cm,
        minimum height=2.2cm,
        rounded corners,
        align=center,
        font=\Large
    },
    exploration/.style={
        rectangle, draw=black, thick,
        minimum width=4.5cm,
        minimum height=2.2cm,
        rounded corners,
        align=center,
        font=\Large,
        text=black
    }
]

\node[layer, label=left:\textbf{Input $\mathbf{x} \in \mathbb{R}^d$}] (input) {
    \textbf{Input Layer}
};

\node[layer, right=of input] (enc1) {
    \textbf{Dense Layer}\\[2pt]
    $\mathbf{h}_1 = \sigma(\mathbf{W}_1 \mathbf{x} + \mathbf{b}_1)$
};
\node[layer, right=of enc1] (enc2) {
    \textbf{Dense Layer}\\[2pt]
    $\mathbf{h}_2 = \sigma(\mathbf{W}_2 \mathbf{h}_1 + \mathbf{b}_2)$
};

\node[latent, right=of enc2, label=below:\textbf{Latent Representation}] (latent) {
    \textbf{Latent Space}\\[2pt]
    $\mathbf{z} = \sigma(\mathbf{W}_3 \mathbf{h}_2 + \mathbf{b}_3)$
};

\node[layer, right=of latent] (dec1) {
    \textbf{Dense Layer}\\[2pt]
    $\mathbf{h}_4 = \sigma(\mathbf{W}_4 \mathbf{z} + \mathbf{b}_4)$
};
\node[layer, right=of dec1] (dec2) {
    \textbf{Dense Layer}\\[2pt]
    $\mathbf{h}_5 = \sigma(\mathbf{W}_5 \mathbf{h}_4 + \mathbf{b}_5)$
};
\node[layer, right=of dec2, label=right:\textbf{Output $\hat{\mathbf{x}}$}] (output) {
    \textbf{Reconstruction}\\[2pt]
    $\hat{\mathbf{x}}$
};

\node[classifier, below=3.2cm of latent, label=below:\textbf{Classifier}] (clf) {
    \textbf{Prediction Layer}\\[2pt]
    $\hat{y} = f_{\text{clf}}(\mathbf{z})$
};
\node[layer, right=of clf, label=right:\textbf{Prediction}] (pred) {
    \textbf{Output Class}\\[2pt]
    $\hat{y}$
};

\node[exploration, left=3.5cm of clf] (explore) {
    \textbf{Latent}\\\textbf{Exploration Layer}
};

\draw[->] (input) -- node[above]{\normalsize DR + BN} (enc1);
\draw[->] (enc1) -- node[above]{\normalsize DR + BN} (enc2);
\draw[->] (enc2) -- node[above]{\normalsize DR + BN} (latent);

\draw[->] (latent) -- node[above]{\normalsize DR + BN} (dec1);
\draw[->] (dec1) -- node[above]{\normalsize DR + BN} (dec2);
\draw[->] (dec2) -- (output);

\draw[->, thick, dashed] (latent.south) -- (clf.north);
\draw[->, thick, dashed] (clf) -- (pred);

\draw[->, thick, dashed] (latent.south west) -- (explore.north east);

\end{tikzpicture}
}
\caption{The CLAIRE architecture. The encoder compresses the input $\mathbf{x} \in \mathbb{R}^d$ into a lower-dimensional latent representation $\mathbf{z} \in \mathbb{R}^k$, passing through multiple dense layers. Dropout Regularization (DR) and Batch Normalization (BN) are applied after each layer to improve generalization and training stability. The decoder reconstructs the input, while the latent code is also used by a downstream classifier to predict the label $\hat{y}$. A dedicated \textbf{Latent Exploration Layer} enables interpretability analysis of $\mathbf{z}$.}
\label{fig:autoencoder_pipeline}
\end{figure*}
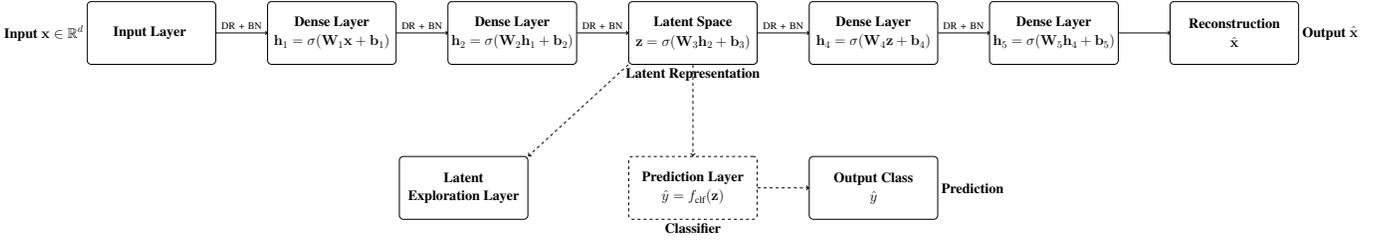

\subsection{Latent Space Construction and Optimization}
This section outlines the steps involved in implementing the model, detailing the architecture of the hybrid approach and the construction of the cost function. 

Denote the training dataset as:
\[
\mathcal{D} = \left\{ (\mathbf{x}^{(i)}, y^{(i)}) \right\}_{i=1}^n, \quad \mathbf{x}^{(i)} \in \mathbb{R}^d, \quad y^{(i)} \in \{0, 1\}
\]
where each $\mathbf{x}^{(i)}$ is a $d$-dimensional feature vector, and $y^{(i)}$ is the corresponding binary label indicating success or failure. The goal is identifying most informative features by training a classifier that maps any new input $\mathbf{x}$ to a predicted label $\hat{y} \in \{0, 1\}$ using an optimized latent space.

The encoding-decoding pipeline is designed to learn an efficient representation of its input data through the process of reconstruction. It is composed of two components, i.e., Encoder and Decoder. The former learns a compressed, latent representation of the input data. The latter attempts to reconstruct the original input from the latent space. We define the mentioned components as follows.

\begin{itemize}
  \item Encoder $f_{\text{enc}}(\cdot; \theta_e): \mathbb{R}^d \rightarrow \mathbb{R}^k$ maps the high-dimensional input $\mathbf{x} \in \mathbb{R}^d$ to a compact latent representation $\mathbf{z} \in \mathbb{R}^k$, where $k \ll d$; and
  \item Decoder $f_{\text{dec}}(\cdot; \theta_d): \mathbb{R}^k \rightarrow \mathbb{R}^d$ attempts to reconstruct the original input from the latent space.
\end{itemize}

Let \(\mathbf{z}^{(i)}\) and \(\hat{\mathbf{x}}^{(i)}\) be the encoding and decoding respectively, i.e.,
\begin{equation}
\mathbf{z}^{(i)} = f_{\text{enc}}(\mathbf{x}^{(i)}; \theta_e), \quad \hat{\mathbf{x}}^{(i)} = f_{\text{dec}}(\mathbf{z}^{(i)}; \theta_d)
\end{equation}
where \(\mathbf{x}^{(i)}\) is the \(i\)-th input data point, \(f_{\text{enc}}\) is the encoding function that maps input \(\mathbf{x}^{(i)}\) to a lower-dimensional latent representation \(\mathbf{z}^{(i)}\), \(\theta_e\) refers to the parameters (weights) of the encoder, which are learned during the training process.
\(\mathbf{z}^{(i)}\) is the compressed latent representation of \(\mathbf{x}^{(i)}\), capturing the essential features of input data. \(\mathbf{z}^{(i)}\) is the latent representation generated by the encoder, and \(f_{\text{dec}}\) is the decoding function that reconstructs the original input from the latent representation \(\mathbf{z}^{(i)}\). \(\theta_d\) refers to the parameters  or weights of the decoder like those in the encoding segment, which are also learned during the training process. \(\hat{\mathbf{x}}^{(i)}\) is the reconstructed version of \(\mathbf{x}^{(i)}\), generated by the decoder.

As mentioned, the encoder transforms \(\mathbf{x}^{(i)}\) into a lower-dimensional latent representation \(\mathbf{z}^{(i)}\), and the decoder attempts to reconstruct the original input from the latent representation. The goal during training is to minimize the reconstruction error between \(\hat{\mathbf{x}}^{(i)}\) and \(\mathbf{x}^{(i)}\), i.e.,
\begin{equation}
\mathcal{L}_{\text{reconstruction}} = \frac{1}{n} \sum_{i=1}^n \left\| \mathbf{x}^{(i)} - f_{\text{dec}}(f_{\text{enc}}(\mathbf{x}^{(i)})) \right\|^2
\end{equation}

The architecture of CLAIRE is illustrated in Fig. \ref{fig:autoencoder_pipeline}. The encoder compresses input $\mathbf{x} \in \mathbb{R}^d$ into a lower-dimensional latent representation $\mathbf{z} \in \mathbb{R}^k$ by passing it through multiple dense layers. The parameters $\theta_e$ and $\theta_d$ of the encoder and decoder, respectively, are optimized during training using gradient descent to minimize the loss function $\mathcal{L}_{\text{DAE}}$ over the dataset. The encoding-decoding pipeline consists of $L$ layers in the encoder and $M$ layers in the decoder. The encoder performs a forward mapping from the input to the latent representation through a sequence of nonlinear transformations:

\begin{equation}
\begin{aligned}
\mathbf{h}^{(1)} &= \sigma(\mathbf{W}^{(1)} \mathbf{x} + \mathbf{b}^{(1)}) \\
\mathbf{h}^{(2)} &= \sigma(\mathbf{W}^{(2)} \mathbf{h}^{(1)} + \mathbf{b}^{(2)}) \\
\rule{0pt}{1.5ex} \vdots \\
\mathbf{z} &= \mathbf{h}^{(L)}
\end{aligned}
\end{equation}

Similarly, the decoder reconstructs the input by reversing the transformation from the latent space:

\begin{equation}
\begin{aligned}
\mathbf{r}^{(1)} &= \sigma(\mathbf{W}^{(1)}_{\text{dec}} \mathbf{z} + \mathbf{b}^{(1)}_{\text{dec}}) \\
\mathbf{r}^{(2)} &= \sigma(\mathbf{W}^{(2)}_{\text{dec}} \mathbf{r}^{(1)} + \mathbf{b}^{(2)}_{\text{dec}}) \\
\rule{0pt}{1.5ex} \vdots \\
\hat{\mathbf{x}} &= \mathbf{r}^{(M)}
\end{aligned}
\end{equation}

Let $\mathcal{L}(\hat{\mathbf{x}}, \mathbf{x})$ denote the reconstruction loss, defined as:

\begin{equation}
\mathcal{L} = \frac{1}{2} \left\| \mathbf{x} - \hat{\mathbf{x}} \right\|^2
\end{equation}
where \(\mathbf{x}\) is the original input and \(\hat{\mathbf{x}}\) is the reconstructed input produced by the decoder. This loss measures the reconstruction error, i.e., how closely the model reproduces its input. The goal during training is to minimize this difference. The factor \(\frac{1}{2}\) is included for mathematical convenience during differentiation.

We now derive the weight updates using the chain rule for vector-valued functions. For any layer $\ell$, the gradient of the loss with respect to weights $\mathbf{W}^{(\ell)}$ is given as:

\begin{equation}
\frac{\partial \mathcal{L}}{\partial \mathbf{W}^{(\ell)}} = \frac{\partial \mathcal{L}}{\partial \mathbf{h}^{(\ell)}} \cdot \frac{\partial \mathbf{h}^{(\ell)}}{\partial \mathbf{W}^{(\ell)}}
\end{equation}

Assuming that each layer consists of a linear transformation followed by a non-linear activation, we define the forward pass for the \(\ell\)-th layer as:

\begin{equation}
\mathbf{h}^{(\ell)} = \sigma(\mathbf{a}^{(\ell)}) \quad \text{where } \mathbf{a}^{(\ell)} = \mathbf{W}^{(\ell)} \mathbf{h}^{(\ell-1)} + \mathbf{b}^{(\ell)}
\end{equation}

Here, \(\mathbf{a}^{(\ell)}\) denotes the pre-activation input to the activation function \(\sigma\), and \(\mathbf{h}^{(\ell-1)}\) is the output of the previous layer. \(\mathbf{W}^{(\ell)}\) and \(\mathbf{b}^{(\ell)}\) are the weight and bias parameters for layer \(\ell\).

Using the chain rule again, the derivative of the activation function with respect to the pre-activation input is:

\begin{equation}
\frac{\partial \mathbf{h}^{(\ell)}}{\partial \mathbf{a}^{(\ell)}} = \sigma'(\mathbf{a}^{(\ell)})
\end{equation}
where \(\sigma'\) denotes the derivative of the activation function with respect to $\mathbf{a}^{(\ell)}$. In this work, we use a combination of leaky ReLU and sigmoid activations.

To update the parameters of the network, we compute the gradient of the loss with respect to the weights as:

\begin{equation}
\frac{\partial \mathcal{L}}{\partial \mathbf{W}^{(\ell)}} = \left( \delta^{(\ell)} \circ \sigma'(\mathbf{a}^{(\ell)}) \right) \cdot \left( \mathbf{h}^{(\ell-1)} \right)^T
\end{equation}
where \(\delta^{(\ell)} = \frac{\partial \mathcal{L}}{\partial \mathbf{h}^{(\ell)}}\) is the error propagated back from the subsequent layer. The symbol \(\circ\) denotes element-wise multiplication.

Similarly, the gradient with respect to the bias is:

\begin{equation}
\frac{\partial \mathcal{L}}{\partial \mathbf{b}^{(\ell)}} = \delta^{(\ell)} \circ \sigma'(\mathbf{a}^{(\ell)})
\end{equation}

To perform optimization, the parameters \(\theta = \{\mathbf{W}, \mathbf{b}\}\) are updated via gradient descent, i.e., $
\theta \leftarrow \theta - \eta \cdot \nabla_{\theta} \mathcal{L},$
where \(\eta > 0\) is the learning rate and \(\nabla_{\theta} \mathcal{L}\) is the gradient of the loss with respect to the parameters.

For the decoder, error propagation begins from the reconstruction loss. The error at the output layer is:

\begin{equation}
\delta^{(M)} = \frac{\partial \mathcal{L}}{\partial \hat{\mathbf{x}}} = \hat{\mathbf{x}} - \mathbf{x}
\end{equation}

This error is then backpropagated layer by layer:

\begin{equation}
\delta^{(M-1)} = \left( \mathbf{W}^{(M)} \right)^T \delta^{(M)} \circ \sigma'(\mathbf{a}^{(M-1)}), \quad \ldots
\end{equation}

Finally, the weights and biases are updated using the computed gradients:

\begin{equation}
\mathbf{W}^{(\ell)} \leftarrow \mathbf{W}^{(\ell)} - \eta \frac{\partial \mathcal{L}}{\partial \mathbf{W}^{(\ell)}}, \quad
\mathbf{b}^{(\ell)} \leftarrow \mathbf{b}^{(\ell)} - \eta \frac{\partial \mathcal{L}}{\partial \mathbf{b}^{(\ell)}}
\end{equation}

\subsubsection{\textbf{Loss Functions}}

As mentioned earlier, CLAIRE aims to minimize the reconstruction error between the input and its reconstructed output, while also regularizing the latent space to ensure compact and meaningful representations. To achieve this, we define two key loss functions: \textit{Reconstruction Loss} and \textit{Latent Variance Loss}, which together form the unsupervised loss term \(\mathcal{L}_{\text{DAE}}\).

\textit{Reconstruction Loss} quantifies the difference between the original input and output of the autoencoder, ensuring that the model effectively captures essential patterns in the data. \textit{Latent Variance Loss} acts as a regularization term that encourages the latent space to remain compact by penalizing excessive variance across the latent dimensions. This discourages the model from relying on irrelevant activations and is particularly important when the latent code is used as input to downstream classifiers. It is defined as:

\begin{equation}
L_{\text{latent}} = \frac{1}{k} \sum_{j=1}^{k} \text{Var}(z_j)
\end{equation}
where \( z_j \) is the \( j \)-th dimension of latent code \( \mathbf{z} \), \( k \) is latent space dimensionality. The loss at this phase is given as:

\begin{equation}
L_{\text{DAE}} = L_{\text{reconstruction}} + \lambda L_{\text{latent}}
\end{equation}
where \( \lambda \) is a hyperparameter that balances reconstruction fidelity and latent space compactness. Minimizing this loss enables CLAIRE to extract representations that are both informative and robust to noise, thus enhancing the performance of the downstream classifier. It is important to note that this loss defines the representation learning objective used in Phase 1 of training. Additional components, including the classification loss and entropy regularization, are also incorporated into the overall optimization objective.

\subsubsection{\textbf{Optimization}}

To ensure stable and efficient training, we adopt a momentum-based optimization approach that dynamically adjusts learning rates. The first and second moments of the gradients are computed as follows:

\begin{equation}
\begin{aligned}
m_t &= \beta_1 m_{t-1} + (1 - \beta_1) \nabla_{\theta} \mathcal{L}, \\
v_t &= \beta_2 v_{t-1} + (1 - \beta_2)(\nabla_{\theta} \mathcal{L})^2 \\
\theta &\leftarrow \theta - \eta \cdot \frac{m_t}{\sqrt{v_t} + \epsilon}
\end{aligned}
\end{equation}
where \(\beta_1\) and \(\beta_2\) are momentum parameters, \(\epsilon\) is a small constant for numerical stability, and \(\eta\) is the learning rate. These updates are applied across all layers of both encoder and decoder until convergence.

To reduce overfitting, we apply dropout regularization by randomly deactivating neurons during training. Let \( \mathbf{h} \in \mathbb{R}^d \) be a layer's activation. The dropout operation is defined as $
\tilde{\mathbf{h}} = \mathbf{m} \circ \mathbf{h},$
where \( \mathbf{m} \in \{0,1\}^d \) is a binary mask sampled from a Bernoulli distribution, i.e.,$
m_i \sim \text{Bernoulli}(p), \quad i = 1, \ldots, d$.

During inference, activations are rescaled by using the dropout rate, i.e., $\mathbf{h}_{\text{test}} = p \cdot \mathbf{h}$. In addition, Batch Normalization (BN) is employed to stabilize training by normalizing activations within each mini-batch. For \( x_i \), the transformation is:

\begin{equation}
\mu_B = \frac{1}{m} \sum_{i=1}^m x_i, \quad \sigma_B^2 = \frac{1}{m} \sum_{i=1}^m (x_i - \mu_B)^2
\end{equation}
\begin{equation}
\hat{x}_i = \frac{x_i - \mu_B}{\sqrt{\sigma_B^2 + \epsilon}}, \quad y_i = \gamma \hat{x}_i + \beta
\end{equation}
where \( m \) is the batch size, \(\gamma\) and \(\beta\) are learnable scale and shift parameters. BN is used to reduce internal covariate shift, improve gradient flow, and enable fast convergence. It is applied after the linear transformation and before the activation function:

\begin{equation}
\mathbf{h}^{(\ell)} = \sigma(\text{BN}(\mathbf{W}^{(\ell)} \mathbf{h}^{(\ell-1)} + \mathbf{b}^{(\ell)}))
\end{equation}

As shown in Fig. \ref{fig:autoencoder_pipeline}, DR and BN are applied after each dense layer in both encoder and decoder. Their integration significantly improves the training stability and generalization performance of the model. This is particularly critical when working with high-dimensional data where overfitting is a common issue.

\subsection{Support Vector Machine with Kernel Trick}

After training, we discard the decoder and retain only the encoder to transform each input $\mathbf{x}^{(i)}$ into its corresponding latent representation $\mathbf{z}^{(i)}$. This results in a transformed dataset:

\begin{equation}
\mathcal{Z} = \left\{ (\mathbf{z}^{(i)}, y^{(i)}) \right\}_{i=1}^n
\end{equation}

We then train a binary Support Vector Machine (SVM) classifier on this lower-dimensional representation. SVM seeks a decision function of the form:

\begin{equation}
f(\mathbf{z}) = \text{sign} \left( \sum_{i=1}^n \alpha_i y^{(i)} K(\mathbf{z}^{(i)}, \mathbf{z}) + b \right)
\end{equation}
where $\alpha_i$ is a dual coefficient, $b$ is a bias term, and $K(\cdot, \cdot)$ is a kernel function satisfying the Mercer’s condition \cite{ZhangShu2018,KangShi2018}.

SVM is trained by solving the following convex optimization problem in its dual form:

\begin{equation}
\begin{aligned}
\min_{\boldsymbol{\alpha}} \quad & \frac{1}{2} \sum_{i=1}^n \sum_{j=1}^n \alpha_i \alpha_j y^{(i)} y^{(j)} K(\mathbf{z}^{(i)}, \mathbf{z}^{(j)}) - \sum_{i=1}^n \alpha_i \\
\text{subject to} \quad & \sum_{i=1}^n \alpha_i y^{(i)} = 0, \quad 0 \leq \alpha_i \leq C, \quad \forall i
\end{aligned}
\end{equation}
where $C > 0$ is a regularization parameter controlling the trade-off between maximizing its margin and minimizing classification error. Note that binary labels are mapped from $\{0,1\}$ to $\{-1,+1\}$ to match the standard SVM formulation, while in neural network training they remain $\{0,1\}$.

For a new and unseen input $\mathbf{x}^*$, the encoder produces its latent representation \(\mathbf{z}^* = f_{\text{enc}}(\mathbf{x}^*)\), and SVM makes a prediction, i.e.,

\begin{equation}
\hat{y} = \text{sign} \left( \sum_{i=1}^n \alpha_i y^{(i)} K(\mathbf{z}^{(i)}, \mathbf{z}^*) + b \right)
\end{equation}

To perform classification in high-dimensional feature space without explicitly computing nonlinear transformations, we utilize the kernel trick. The kernel function implicitly defines an inner product in the transformed space:

\begin{equation}
K(\mathbf{z}^{(i)}, \mathbf{z}^{(j)}) = \langle \phi(\mathbf{z}^{(i)}), \phi(\mathbf{z}^{(j)}) \rangle
\end{equation}
where \( \phi(\cdot) \) is an implicit nonlinear mapping, and \( K(\cdot, \cdot) \) is a valid Mercer kernel. We conduct a comparative evaluation of four widely used kernels:

\begin{enumerate}
    \item \textbf{Linear kernel:} \( K(\mathbf{z}, \mathbf{z}') = \mathbf{z}^\top \mathbf{z}' \). It is computationally efficient but limited to linear separability.
    \item \textbf{Polynomial kernel:} \( K(\mathbf{z}, \mathbf{z}') = (\mathbf{z}^\top \mathbf{z}' + c)^d \). It captures nonlinear interactions but is sensitive to feature scaling and prone to overfitting in high-dimensional space.
    \item \textbf{Radial Basis Function (RBF) kernel:} \( K(\mathbf{z}, \mathbf{z}') = \exp(-\gamma \|\mathbf{z} -\mathbf{z}'\|^2) \). It provides flexible, localized decision boundaries; robust in high-dimensional settings.
    \item \textbf{Sigmoid kernel:} \( K(\mathbf{z}, \mathbf{z}') = \tanh(\alpha \mathbf{z}^\top \mathbf{z}' + \beta) \). It is inspired by neural networks but occasionally suffers from poor convergence.
\end{enumerate}

Among these, the RBF kernel demonstrates superior performance on the learned latent representations due to its ability to model complex, nonlinear decision boundaries. Based on our empirical findings, we thus select the RBF kernel for the final SVM classifier. Algorithm~\ref{alg:dae_svm} realizes the full two-phase training procedure of the proposed CLAIRE framework.

\begin{algorithm}[htbp]
\caption{Two-Phase CLAIRE Framework}
\label{alg:dae_svm}
\begin{algorithmic}[1]
\Require Training data $\mathcal{D} = \{(\mathbf{x}^{(i)}, y^{(i)})\}_{i=1}^n$
\Statex \hspace{\algorithmicindent} where $\mathbf{x}^{(i)} \in \mathbb{R}^d$, $y^{(i)} \in \{0, 1\}$
\Ensure Predictive model for $\hat{y}$ from input $\mathbf{x}$

\State \textbf{Initialize:} Encoder parameters $\theta_e$, Decoder parameters $\theta_d$
\State \textbf{Define:} Encoder $f_{\text{enc}}(\mathbf{x}; \theta_e)$, Decoder $f_{\text{dec}}(\mathbf{z}; \theta_d)$
\State \textbf{Initialize}: Classifier parameters $\theta_c$
\State \textbf{Define}: Classifier $f_{\text{clf}}(\mathbf{z};\theta_c)$ and $\hat{y}^{(i)} = f_{\text{clf}}(z^{(i)};\theta_c)$
\State \textbf{Define:} Corruption function $\mathcal{C}(\cdot)$

\State \textbf{Define:} Reconstruction loss:
\Statex \hspace{\algorithmicindent} $\mathcal{L}_{\text{recon}} = \dfrac{1}{n} \sum\limits_{i=1}^{n} \left\| \mathbf{x}^{(i)} - f_{\text{dec}}(f_{\text{enc}}(\tilde{\mathbf{x}}^{(i)})) \right\|^2$
\Statex \hspace{\algorithmicindent} where $\tilde{\mathbf{x}}^{(i)} = \mathcal{C}(\mathbf{x}^{(i)})$

\State \textbf{Define:} Latent variance loss:
\Statex \hspace{\algorithmicindent} $\mathcal{L}_{\text{latent}} = \dfrac{1}{d} \sum\limits_{j=1}^{d} \text{Var}(z_j)$
\Statex \hspace{\algorithmicindent} where $z_j$ is the $j$-th latent unit

\State \textbf{Define:} Representation learning loss:
\Statex \hspace{\algorithmicindent} $\mathcal{L}_{\text{DAE}} = \mathcal{L}_{\text{recon}} + \lambda \mathcal{L}_{\text{latent}}$
\Statex \hspace{\algorithmicindent} where $\lambda > 0$ is a regularization weight

\State \textbf{Define:} Supervised classification loss:
\Statex \hspace{\algorithmicindent} $\mathcal{L}_{\text{clf}} = - \dfrac{1}{n} \sum\limits_{i=1}^{n} \Big[ y^{(i)} \log \hat{y}^{(i)} + (1 - y^{(i)}) \log (1 - \hat{y}^{(i)}) \Big]$

\State \textbf{Define:} Entropy regularization:
\Statex \hspace{\algorithmicindent}
$\mathcal{L}_{\text{ent}} 
= - \dfrac{1}{n} 
\sum\limits_{i=1}^{n}
\Big[
\hat{y}^{(i)} \log \hat{y}^{(i)} 
+ (1 - \hat{y}^{(i)}) \log (1 - \hat{y}^{(i)})
\Big]$

\State \textbf{Define:} Total training objective:
\Statex \hspace{\algorithmicindent} $\mathcal{L}_{\text{Total}} = \mathcal{L}_{\text{DAE}} + \alpha \mathcal{L}_{\text{clf}} + \beta \mathcal{L}_{\text{ent}}$
\Statex \hspace{\algorithmicindent} where $\alpha, \beta > 0$ are weighting coefficients

\State \textbf{Define:} Dropout rate $p$ and BatchNorm function $\mathcal{B}(\cdot)$

\vspace{0.3em}
\Statex \textit{/* Phase 1: Optimized Encoder-Decoder and Joint Representation */}
\For{$\text{epoch} = 1$ to $N$}
    \ForAll{mini-batches $\{\mathbf{x}^{(i)}\}$}
        \State Corrupt input: $\tilde{\mathbf{x}}^{(i)} \gets \mathcal{C}(\mathbf{x}^{(i)})$
        \State Encode: $\mathbf{z}^{(i)} \gets f_{\text{enc}}(\tilde{\mathbf{x}}^{(i)}; \theta_e)$
        \State Apply dropout: $\mathbf{z}^{(i)} \gets \text{Dropout}(\mathbf{z}^{(i)}, p)$
        \State Apply BN: $\mathbf{z}^{(i)} \gets \mathcal{B}(\mathbf{z}^{(i)})$
        \State Decode: $\hat{\mathbf{x}}^{(i)} \gets f_{\text{dec}}(\mathbf{z}^{(i)}; \theta_d)$
        \State Compute total loss $\mathcal{L}_{\text{Total}}$
        \State Update $\theta_e$, $\theta_d, \theta_c$ via gradient descent
    \EndFor
\EndFor

\vspace{0.3em}
\Statex \textit{/* Phase 2: Training an SVM on Frozen Latent Representations; */}
\State Generate latent dataset: $\mathcal{Z} = \{(f_{\text{enc}}(\mathbf{x}^{(i)}), y^{(i)})\}_{i=1}^n$
\State Train SVM by solving:
\Statex \hspace{\algorithmicindent} $\min_{\boldsymbol{\alpha}} \frac{1}{2} \sum_{i,j} \alpha_i \alpha_j y^{(i)} y^{(j)} K(\mathbf{z}^{(i)}, \mathbf{z}^{(j)}) - \sum_i \alpha_i$
\Statex \hspace{\algorithmicindent} subject to $\sum_i \alpha_i y^{(i)} = 0,\quad 0 \leq \alpha_i \leq C$

\vspace{0.3em}
\Statex \textit{/* Inference */}
\State \textbf{Given:} new test input $\mathbf{x}^*$
\State Compute latent code: $\mathbf{z}^* = f_{\text{enc}}(\mathbf{x}^*)$
\State Predict label:
\Statex \hspace{\algorithmicindent} $\hat{y} = \text{sign} \left( \sum_{i=1}^n \alpha_i y^{(i)} K(\mathbf{z}^{(i)}, \mathbf{z}^*) + b \right)$

\end{algorithmic}
\end{algorithm}

\subsection{Interpreting Latent Representations}

To improve the transparency of CLAIRE and explain the internal decision logic, we introduce a post hoc interpretability phase based on a game-theoretic approach. Specifically, we apply SHapley Additive exPlanations to encoder to assess how individual input features influence latent space. This process enables us to analyze which features drive the learned latent representations. 

\begin{algorithm}[htbp]
\caption{Latent Space Interpretation}
\label{alg:shap_explainer}
\begin{algorithmic}[1]
\Require Trained CLAIRE model, input data $\mathbf{X}_{\text{train}}, \mathbf{X}_{\text{test}}$
\Ensure SHAP values explaining feature contributions to latent dimensions

\State \textbf{Step 1:} Wrap encoder into a standalone model
\[
\text{Define: } f_{\text{enc}}(\mathbf{x}) \gets \text{Encoder module from trained CLAIRE}
\]
\State \textbf{Step 2:} Select representative background samples
\[
\mathbf{X}_{\text{bg}} \gets \text{First } k \text{ samples from } \mathbf{X}_{\text{train}}
\]

\State \textbf{Step 3:} Define SHAP masker
\[
\text{masker} \gets \texttt{shap.maskers.Independent}(\mathbf{X}_{\text{bg}})
\]

\State \textbf{Step 4:} Create SHAP explainer
\[
\text{explainer} \gets \texttt{shap.Explainer}(f_{\text{enc}}, \text{masker})
\]

\State \textbf{Step 5:} Compute SHAP values for test data
\[
\mathbf{X}_{\text{test}}' \gets \text{First } m \text{ samples from } \mathbf{X}_{\text{test}}
\]
\[
\text{shap\_values} \gets \text{explainer}(\mathbf{X}_{\text{test}}')
\]

\State \textbf{Step 6:} Visualize beeswarm plots for selected latent dimensions
\[
\texttt{shap.plots.beeswarm}(\text{shap\_values}[:, :, i])
\]
\Statex \hspace{\algorithmicindent} \texttt{i = 1 \dots 4}

\end{algorithmic}
\end{algorithm}

Algorithm \ref{alg:shap_explainer} computes SHAP values for each latent dimension of encoder. The SHAP plots reveal which input features most strongly influence the learned latent space across the input. Such interpretability is essential in real-world industrial settings, where understanding feature importance can inform root cause analysis, domain expert validation, and regulatory compliance.

\section{Experimental Results}\label{section.Results}
This section presents the experimental results obtained from training and evaluating the proposed CLAIRE. We investigate the influence of several hyperparameters, such as the number of neurons and layers, mini-batch size, learning rate, and loss function components, on model performance. In addition, we compare CLAIRE’s classification performance with a traditional baseline and analyze the learned latent representations.

\subsection{Effect of Architectural and Training Choices}
The architecture plays a crucial role in learning meaningful feature representations. We experiment with different layer and neuron counts, ultimately selecting an encoder with two layers of 128 and 64 neurons, mirrored by the decoder. Increasing the number of neurons tends to improve both reconstruction and classification accuracy by enabling the model to capture more complex patterns. However, deeper models exhibit the signs of overfitting, as reflected in the increased variance of loss values across epochs. Conversely, reducing neurons leads to underfitting where the model fails to capture key characteristics of the data. These results underscore the importance of architectural balance. In addition, we observe that excessively deep or wide architectures tend to destabilize the latent variance regularization term, leading to less compact embeddings despite improved reconstruction, further motivating a balanced architectural design. We evaluate mini-batch sizes of 32 and 64. The former introduces higher gradient variance, resulting in noisier updates and slower convergence. The latter requires much more memory than the former. It improves convergence stability and training efficiency. Therefore, we adopt a mini-batch size of 64. This choice also contributes to more stable optimization of the latent variance regularization by reducing stochastic fluctuations in latent statistics across batches. 
We emphasize that the choice of mini-batch sizes is guided by practical considerations commonly adopted in industrial deep learning applications. Very small batch sizes tend to introduce excessive gradient noise and unstable latent statistics, while substantially larger batches offer diminishing returns in convergence quality and incur unnecessary memory and computational overhead. Preliminary trials with larger batch sizes did not yield measurable improvements in reconstruction accuracy, classification performance, or latent compactness. Consequently, the selected range represents a balanced and practically sufficient operating regime rather than an exhaustive hyperparameter search. For optimization, we test the learning rates of \(1 \times 10^{-2}\) and \(1 \times 10^{-3}\). The latter provides smooth convergence with consistent loss reduction, whereas the former introduces instability despite faster early training. As a result, \(1 \times 10^{-3}\) is selected for final training. Overall, these architectural and training choices jointly ensure stable convergence behavior while preserving the discriminative and compact structure of the learned latent space.

\subsection{Loss Function Behavior and Regularization}
\begin{figure}[!t]
    \centering
    \includegraphics[width=0.90\linewidth]{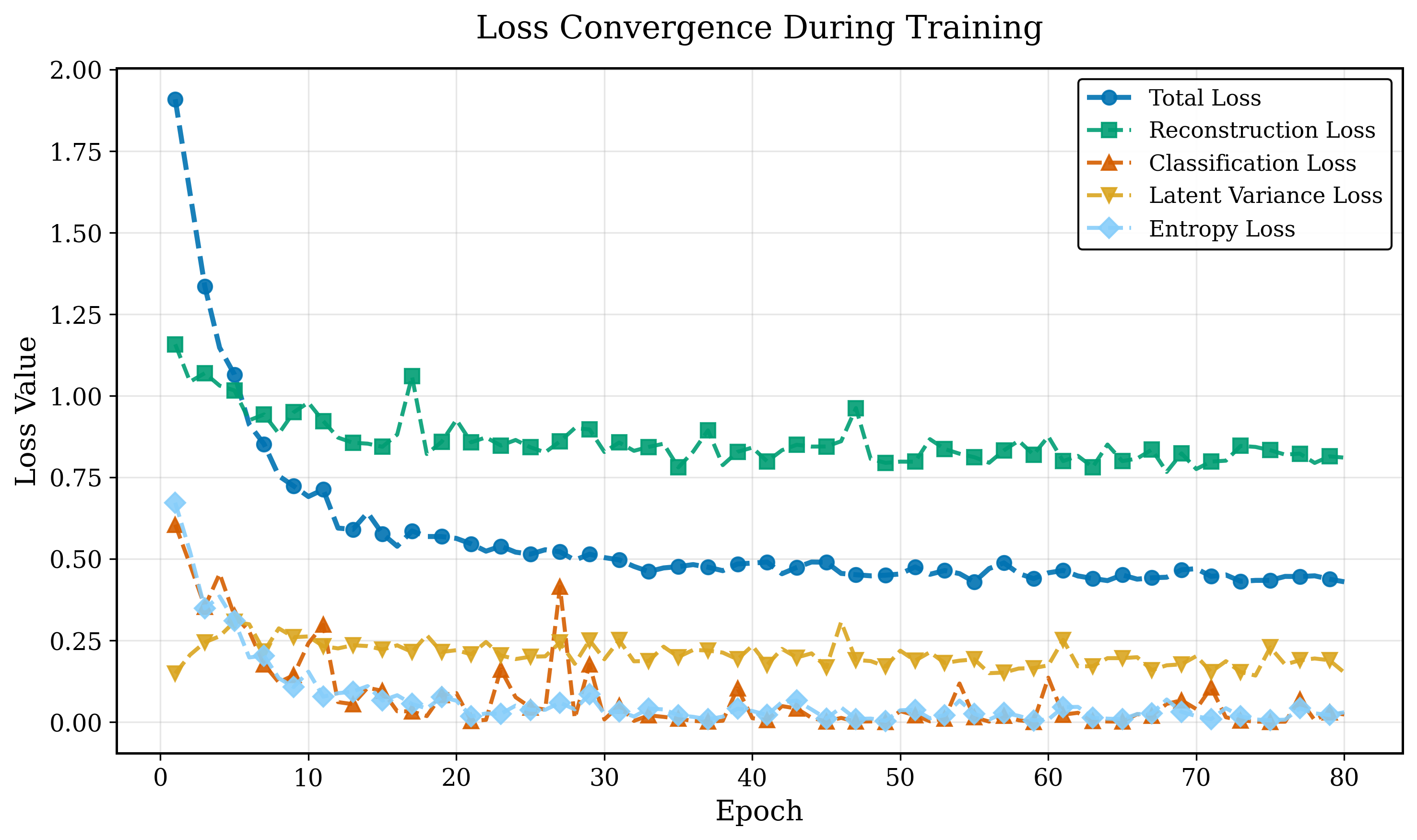}
    \caption{Loss convergence over training epochs on the SECOM dataset.}
    \label{fig:loss_convergence}
\end{figure}

CLAIRE's objective combines multiple losses: reconstruction, classification, latent variance regularization, and entropy regularization. The reconstruction loss ensures faithful input reconstruction, while the classification one improves prediction performance. Latent variance loss encourages compact representations. The entropy regularization loss prevents overconfident predictions. 
From an optimization perspective, the latent variance regularization acts as a geometry-shaping constraint that gradually reduces the dispersion of latent embeddings without dominating the reconstruction objective. We apply Dropout Regularization (rate 0.3) and Batch Normalization (momentum 0.9) after each dense layer to improve generalization and stability. These regularization mechanisms further mitigate overfitting and help stabilize the interaction between reconstruction and latent compactness objectives during training. Fig.~\ref{fig:loss_convergence} shows the convergence curves across 80 training epochs on the SECOM dataset. The representation learning loss, the classification loss, and entropy-based regularization are plotted  in their non-weighted forms to illustrate convergence behavior, while the total loss reflects their weighted aggregation during training. Due to loss weighting, normalization effects (Batch Normalization), and Dropout Regularization, the total loss magnitude is not directly comparable to the standalone terms. The total loss decreases rapidly during the first 30 epochs and then stabilizes. Classification and entropy regularization losses converge after around 20--30 epochs, while the reconstruction and latent variance losses converge early. Notably, the early stabilization of the latent variance loss indicates that compact latent structures are established in the initial training phase and remain stable thereafter. These trends suggest that meaningful convergence is achieved within 40 epochs, allowing us to reduce training time without performance loss. Consequently, we fix the number of epochs at 40 in the final implementation. For the TEP dataset, we keep all hyperparameters approximately the same, except for reducing the number of epochs to 30 due to its smaller number of features and faster convergence. This consistency across datasets further confirms the robustness of the proposed training strategy.

\subsection{Comparative Evaluation}
We have benchmarked the proposed model against several baseline approaches to assess its effectiveness. In addition to a traditional \textit{Support Vector Machine (SVM)} with an \textit{RBF kernel} trained directly on raw features, we have also included representation-learning-based models, i.e., a standard \textit{Autoencoder (AE)}, a \textit{Variational Autoencoder (VAE)}, and a \textit{$\beta$-VAE}, trained on both SECOM and TEP. These models are selected as baselines because they are capable of mapping the input features into lower-dimensional latent spaces, which allows a fair comparison with our proposed method that also relies on learning discriminative latent representations.

Each of these baselines have been implemented by using the same data preprocessing pipeline, initialization strategy, and training protocol to eliminate biases introduced by differing experimental conditions. In addition, hyperparameters such as learning rate, batch size, latent dimensionality, and dropout rate have been tuned individually for each model through empirical validation. This ensures that all baselines operate under optimized yet comparable conditions. The AE serves as a deterministic reconstruction-based baseline, while the VAE introduces probabilistic latent variable modeling with Kullback–Leibler (KL) divergence regularization. The $\beta$-VAE has extended the VAE framework by scaling the KL term with a factor $\beta > 1$, thereby encouraging more disentangled latent representations. By including these variants, the evaluation captures both deterministic and probabilistic reconstruction paradigms, as well as a disentanglement-oriented extension.

Overall, the design and tuning of these baselines provide a robust foundation for evaluation, ensuring that performance improvements observed in the proposed model are attributable to its methodological contributions. Table~\ref{tab:comparison_results} shows that CLAIRE significantly outperforms all baselines in both accuracy and F1 score, highlighting the advantage of its learned latent features in capturing class-discriminative structure.

\begin{table}[!t]
\centering
\caption{Performance comparison of all baseline models and the proposed method (CLAIRE) on the SECOM and TEP datasets.}
\label{tab:comparison_results}
\footnotesize
\begin{tabular}{lcc}
\toprule
\multicolumn{3}{c}{\textbf{SECOM Dataset}} \\
\midrule
\textbf{Model} & \textbf{Accuracy} & \textbf{F1 Score} \\
\midrule
SVM (RBF kernel)& 0.84 & 0.84 \\
AE              & 0.86 & 0.86 \\
VAE              & 0.85 & 0.84 \\
$\beta$-VAE      & 0.83 & 0.82 \\
CLAIRE (Proposed) & 0.94 & 0.93 \\
\midrule
\multicolumn{3}{c}{\textbf{TEP Dataset}} \\
\midrule
\textbf{Model} & \textbf{Accuracy} & \textbf{F1 Score} \\
\midrule
SVM (RBF kernel) & 0.84 & 0.82 \\
AE               & 0.84 & 0.84 \\
VAE              & 0.82 & 0.83 \\
$\beta$-VAE      & 0.86 & 0.86 \\
CLAIRE (Proposed) & 0.92 & 0.92 \\
\bottomrule
\end{tabular}
\end{table}

\subsection{Latent Space Visualization}
\begin{figure*}[!t]
    \centering
    \includegraphics[width=\textwidth]{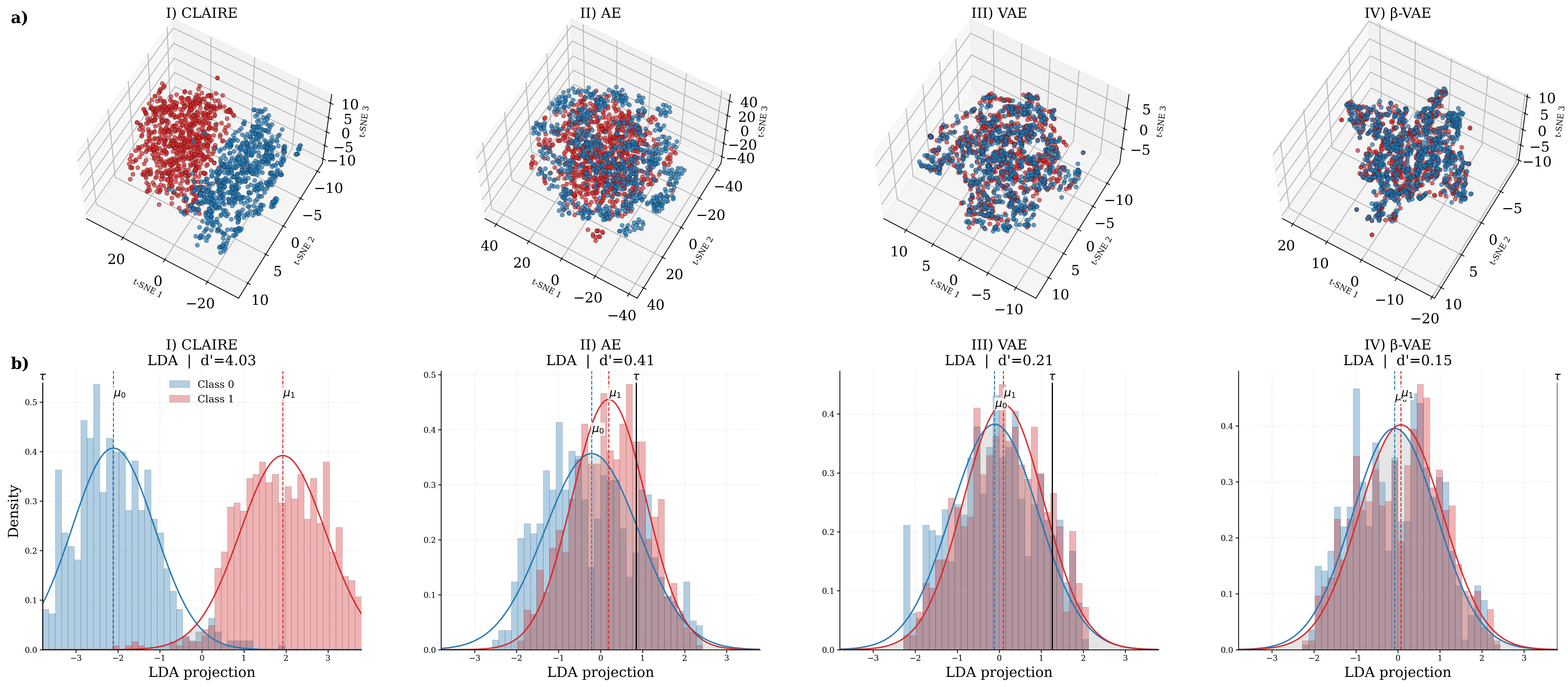}
    \caption{
    Visualization of latent representations for SECOM. 
    a) Three-dimensional t-SNE embeddings for four models: (I) CLAIRE, (II) Standard Autoencoder (AE), (III) Variational Autoencoder (VAE), and (IV) $\beta$-VAE. 
    Each point represents a sample colored by its class label (blue: Class~0, red: Class~1). 
    b) LDA one-dimensional projections of latent spaces. 
    Histograms and Gaussian fits illustrate the class distributions, with vertical dashed lines marking class means ($\mu_0$, $\mu_1$) and solid black lines indicating the discriminant threshold $\tau$. 
    The reported $d'$ values quantify the separation between classes, highlighting the superior discriminability of CLAIRE compared to baseline models.}
    \label{fig:tsne_latent}
\end{figure*}

\begin{figure*}[t]
    \centering
    \includegraphics[width=\textwidth]{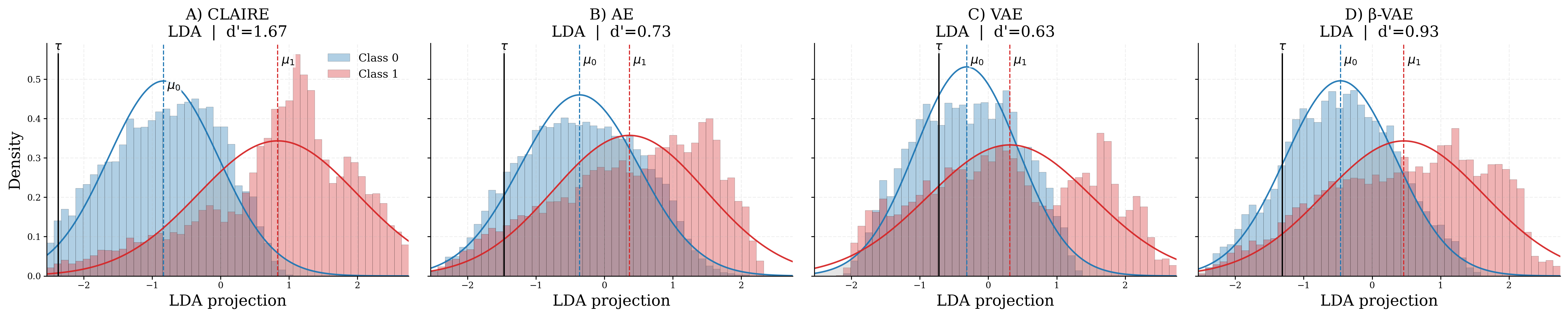}
    \caption{
    LDA one-dimensional projections of the latent representations obtained for TEP. CLAIRE demonstrates a substantially larger separation compared to the baseline models.
    }
    \label{fig:lda_tep}
\end{figure*}

To further assess the learned latent space, we apply \textit{t}-distributed Stochastic Neighbor Embedding (\textit{t}-SNE) to project the high-dimensional latent codes into 3D for visualization. Fig.~\ref{fig:tsne_latent}(a) provides a comparative visualization of the latent spaces learned by CLAIRE and three baseline models (AE, VAE, and $\beta$-VAE) on SECOM (samples are color-coded by class label). While the baseline models produce largely overlapping representations, CLAIRE exhibits a clearer spatial separation between the two classes. 
Fig.~\ref{fig:tsne_latent}(b) presents the Linear Discriminant Analysis (LDA) projections, together with Gaussian density estimates. The class means ($\mu_0$, $\mu_1$), discriminant threshold $\tau$, and the separability index $d'$ are also shown. 
The results reveal that CLAIRE achieves the highest $d'$ value ($d'=4.03$), indicating significantly stronger class discrimination, whereas the baseline models demonstrate substantially weaker separation (with their $d'<0.5$). This confirms the advantage of CLAIRE in producing more structured and discriminative latent representations. The resulting clusters are well-separated, with minimal overlap between two classes, indicating that CLAIRE learns highly discriminative latent representations. This clear separation supports the impact of latent variance regularization in promoting compact, class-specific encoding. The latent structure not only improves reconstruction and classification, but also reveals internal structure aligned with decision boundaries. Similarly, for \textit{TEP}, the LDA projections of the learned latent features (Fig.~\ref{fig:lda_tep}) also show clear separation between classes. CLAIRE achieves a substantially higher separability index than all baseline models, indicating that the same discriminative latent representation is consistently observed on this dataset as well.

\subsection{Explaining Latent Representations via a Game-Theoretic Technique}
To interpret the internal behavior of CLAIRE, we apply a game-theory-based interpretability technique by using SHapley Additive exPlanations (SHAP). Given that the encoder is responsible for transforming high-dimensional sensor data into a compact latent representation, we isolate the encoder subnetwork and treat it as a standalone model. Using SHAP, we compute how each input feature contributes to variations in each latent dimension. A representative subset of the training data is used as the background distribution. SHAP values are then computed on test inputs to assess the marginal contribution of each input feature to the encoder’s output. This approach allows us to understand how perturbations in raw input data affect the learned latent representations, offering a window into the model’s internal structure. The importance of input features in SHAP analysis is conventionally quantified by the mean absolute SHAP value across all samples, defined as
\[
\text{Importance}(f_i) = \frac{1}{N} \sum_{j=1}^{N} \big| \hat{S}_{i,j} \big|,
\]
where $\hat{S}_{i,j}$ denotes the SHAP attribution of feature $f_i$ for sample $j$. 
A higher value of this metric indicates that the corresponding feature exerts a stronger influence on the model’s output (in this case, the latent representation or the classification outcome). For SECOM, Fig.~\ref{fig:shap_column} shows SHAP plots for four selected latent dimensions. Each point represents a feature's SHAP value for an individual sample. It is important to note that the sign of SHAP values provides additional interpretability, i.e., positive attributions (depicted in red) indicate that the feature increases the latent dimension’s activation for a given instance, whereas negative ones (depicted in blue) indicate that the feature decreases it. Hence, the color distribution across samples reflects the direction of feature contributions, while the magnitude of the mean absolute SHAP values is the primary determinant of global feature importance. These plots reveal which input features have the greatest influence on each latent unit. For example, in Fig.~\ref{fig:shap_column}(a), Features 13, 5, and 4 are the most influential contributors to Latent Dimension 0. In Fig.~\ref{fig:shap_column}(b), Feature 13 stands out again. Together with Features 5 and 61, they play a significant role in Dimension 5. In Dimension 10 (Fig.~\ref{fig:shap_column}(c)), Features 13, 24, and 15 are the top contributors. Feature 13 again emerges as influential in Dimension 15, alongside Features 12, 88, and 80 (Fig.~\ref{fig:shap_column}(d)). To assess global feature importance, we have aggregated the mean absolute SHAP values across all latent dimensions and test samples. Table~\ref{tab:shap-ranking} lists the top 10 input features ranked by their average contribution to the latent space. These globally dominant features consistently influence multiple latent dimensions, underscoring their critical role in CLAIRE’s representation learning process. We use 64 latent dimensions for this analysis on SECOM due to its large number of input features, while for TEP, the same analysis is performed with 32 latent dimensions given its smaller feature set. Table~\ref{tab:tepshap-ranking} lists the top 10 input features for TEP.

\begin{figure}[htbp]
    \centering
    \captionsetup{font=small}

    \foreach \i/\name in {1/Latent Dimension 0, 2/Latent Dimension 5, 3/Latent Dimension 10, 4/Latent Dimension 15} {
        \begin{subfigure}{\columnwidth}
            \centering
            \includegraphics[width=0.90\columnwidth]{figures/\i.png}
            \caption{\name}
        \end{subfigure}
        \par\vspace{0.3em}
    }

    \caption{SHAP plots for the SECOM dataset, illustrating the most influential input features across four selected latent dimensions.}
    \label{fig:shap_column}
\end{figure}

\begin{table}[ht]
    \centering
    \scriptsize
    \caption{Top 10 globally important input features for the SECOM dataset, ranked according to their mean SHAP values (MSV) across all 64 latent dimensions.}
    \label{tab:shap-ranking}
    \renewcommand{\arraystretch}{1.2}
    \setlength{\tabcolsep}{4pt}
    \begin{tabular}{c|c|P{1.3cm}|P{1.3cm}|c|P{1.3cm}|P{1.1cm}}
        \toprule
        \textbf{Rank} & \textbf{Index} &
        \makecell{\textbf{SHAP}\\\textbf{Dim. 0}} &
        \makecell{\textbf{SHAP}\\\textbf{Dim. 1}} &
        \textbf{\dots} &
        \makecell{\textbf{SHAP}\\\textbf{Dim. 64}} &
        \makecell{\textbf{MSV}} \\
        \midrule
        1 & Feature 13 & 0.0801 & 0.0283 & \dots & 0.0302 & 0.0380 \\
        2 & Feature 5 & 0.0712 & 0.0558 & \dots & 0.0310 & 0.0368 \\
        3 & Feature 24 & 0.0510 & 0.0212 & \dots & 0.0291 & 0.0356 \\
        4 & Feature 12 & 0.0604 & 0.0273 & \dots & 0.0306 & 0.0286 \\
        5 & Feature 15 & 0.0583 & 0.0241 & \dots & 0.0148 & 0.0282 \\
        6 & Feature 39 & 0.0470 & 0.0207 & \dots & 0.0201 & 0.0267 \\
        7 & Feature 88 & 0.0562 & 0.0164 & \dots & 0.0222 & 0.0264 \\
        8 & Feature 48 & 0.0452 & 0.0162 & \dots & 0.0092 & 0.0262 \\
        9 & Feature 21 & 0.0360 & 0.0266 & \dots & 0.0128 & 0.0260 \\
        10 & Feature 80 & 0.0241 & 0.0141 & \dots & 0.0201 & 0.0241 \\
        \bottomrule
    \end{tabular}
\end{table}

\begin{figure}[htbp]
    \centering
    \includegraphics[width=0.90\linewidth]{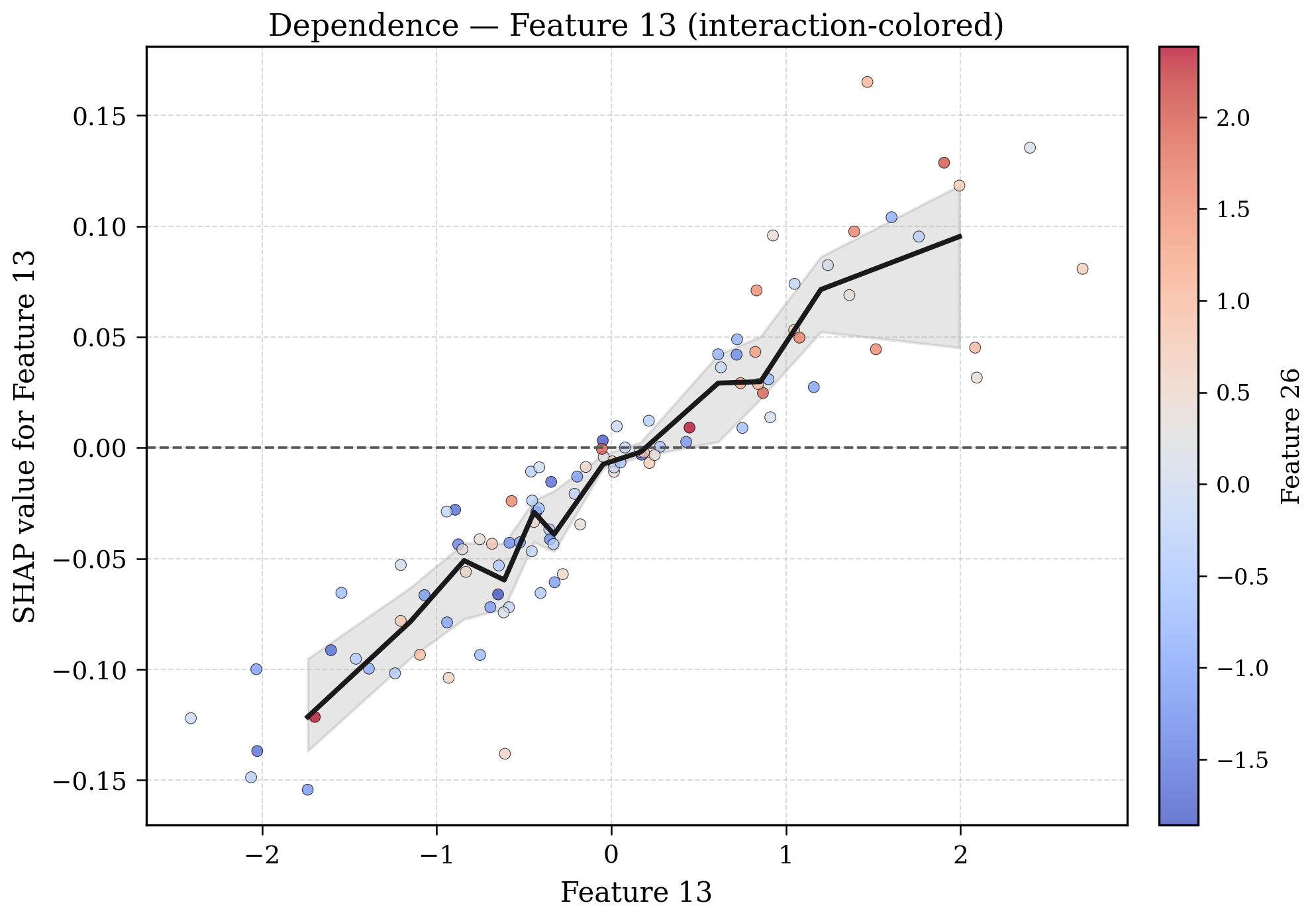}
    \caption{SHAP dependence plot for Feature 13 on the SECOM dataset, with color coding based on Feature 26. The plot demonstrates how changes in Feature 13 influence the model’s latent representation, while also revealing interaction effects that are modulated by Feature 26.}
    \label{fig:shap_dependence}
\end{figure}

\begin{figure}[htbp]
    \centering
    \includegraphics[width=0.90\linewidth]{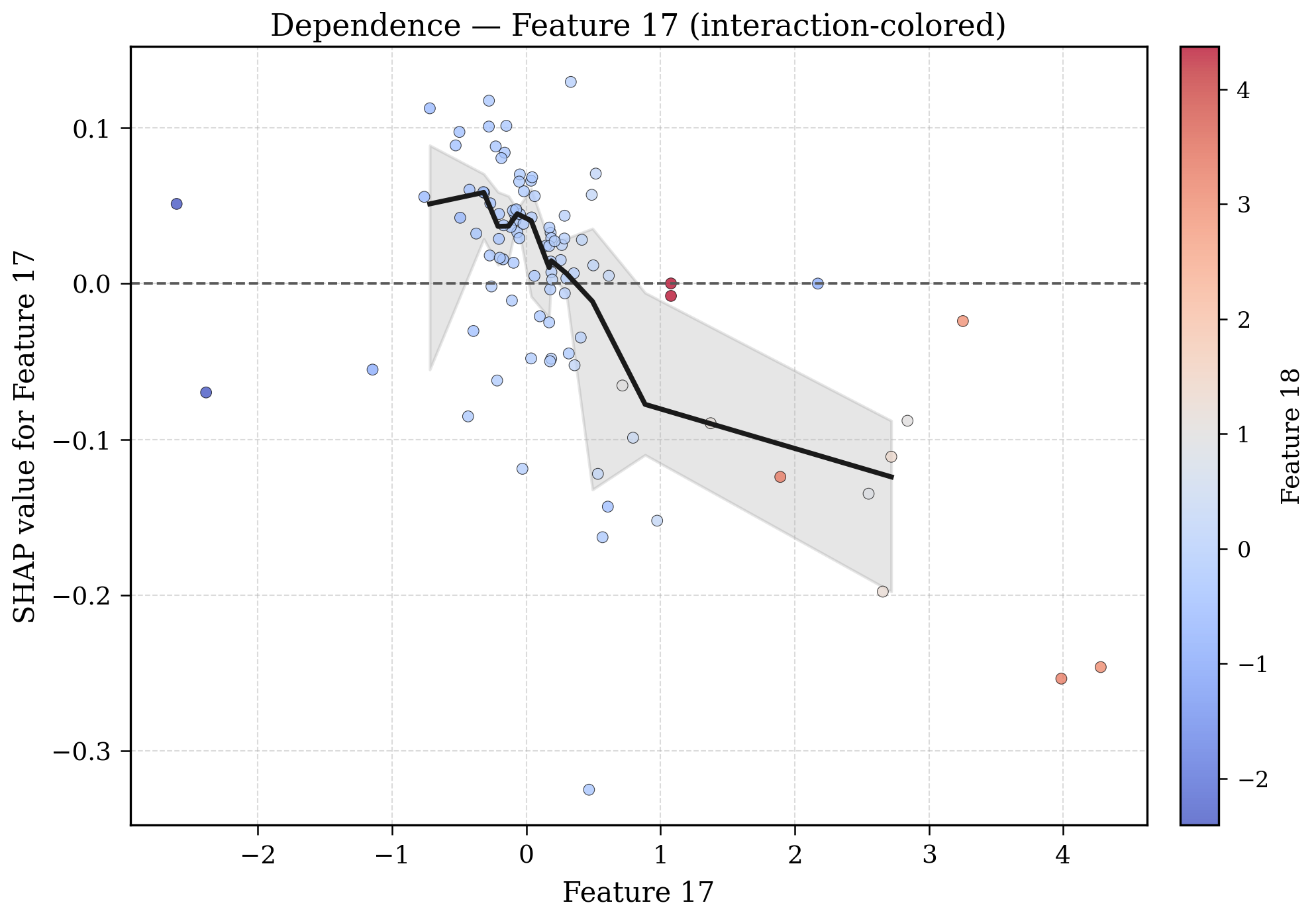}
    \caption{SHAP dependence plot for Feature 17 on the TEP dataset, with color coding by Feature 18. The figure highlights interaction effects that are further modulated by Feature 18.}
    \label{fig:shap_dependence_tep}
\end{figure}

\begin{table}[ht]
    \centering
    \scriptsize
    \caption{Top 10 globally important input features for the TEP dataset, ranked based on their MSVs across all 32 latent dimensions.}
    \label{tab:tepshap-ranking}
    \renewcommand{\arraystretch}{1.2}
    \setlength{\tabcolsep}{4pt}
    \begin{tabular}{c|c|P{1.3cm}|P{1.3cm}|c|P{1.3cm}|P{1.1cm}}
        \toprule
        \textbf{Rank} & \textbf{Index} &
        \makecell{\textbf{SHAP}\\\textbf{Dim. 0}} &
        \makecell{\textbf{SHAP}\\\textbf{Dim. 1}} &
        \textbf{\dots} &
        \makecell{\textbf{SHAP}\\\textbf{Dim. 32}} &
        \makecell{\textbf{MSV}} \\
        \midrule
        1 & Feature 17 & 0.0593 & 0.1040 & \dots & 0.1002 & 0.1004 \\
        2 & Feature 49 & 0.0543 & 0.0768 & \dots & 0.0746 & 0.0752 \\
        3 & Feature 51 & 0.0821 & 0.0614 & \dots & 0.0622 & 0.0641 \\
        4 & Feature 50 & 0.0422 & 0.0483 & \dots & 0.0510 & 0.0479 \\
        5 & Feature 12 & 0.0272 & 0.0414 & \dots & 0.0415 & 0.0424 \\
        6 & Feature 16 & 0.0796 & 0.0409 & \dots & 0.0401 & 0.0417 \\
        7 & Feature 6 & 0.0339 & 0.0320 & \dots & 0.0309 & 0.0314 \\
        8 & Feature 45 & 0.0228 & 0.0281 & \dots & 0.0280 & 0.0299 \\
        9 & Feature 10 & 0.0203 & 0.0290 & \dots & 0.0291 & 0.0283 \\
        10 & Feature 43 & 0.0213 & 0.0278 & \dots & 0.0247 & 0.0264 \\
        \bottomrule
    \end{tabular}
\end{table}
We next identify the true root causes of faults by finding which input features are statistically responsible for failure instances (Class 1). To do so, we extract importance values specifically for predictions labeled as failures. This allows us to identify features that exhibit substantially higher SHAP values in failure samples (Class 1) compared to successful samples (Class 0). We have found that Features 13, 24, and 12, consistently show high SHAP values for faulty predictions on SECOM. Beyond individual feature contributions, SHAP dependence plots enable the analysis of feature interactions, where the effect of one feature is conditioned on the value of another. Fig.~\ref{fig:shap_dependence} presents a SHAP dependence plot for Feature 13, one of the most influential variables in predicting failure instances, showing that its SHAP values generally increase with larger feature values, suggesting a stronger positive contribution to latent space. Specifically, instances with high values of Feature 26 (depicted in red) display predominantly positive SHAP values for Feature 13, whereas instances with low values of Feature 26 (depicted in blue) exhibit neutral or negative SHAP contributions. This observation indicates that the impact of Feature 13 is not independent, but rather strongly modulated by Feature 26. A positive SHAP value suggests that higher readings of Feature 13 increase the likelihood of the product being classified as faulty. The color gradient corresponds to the values of Feature 26, revealing a strong interaction effect: the impact of Feature 13 intensifies when Feature 26 is also high. Such behavior suggests a \textbf{compound effect} where co-occurring elevated readings from both features may reflect underlying failure conditions in the manufacturing process. 
For TEP, Feature 17 is identified as the most influential variable for predicting Class 1 instances. The dependence plot in Fig.~\ref{fig:shap_dependence_tep} illustrates the relationship between this feature and its SHAP values, with color coding reflecting the values of Feature 18. The color distribution highlights a strong interaction with Feature 18. Samples with low values of Feature 18 (depicted in blue) exhibit more negative SHAP contributions from Feature 17, whereas samples with high values of Feature 18 (depicted in red) show less pronounced decreases. This observation indicates that the effect of Feature 17 is not independent but modulated by Feature 18, thereby exemplifying a compound effect between the two variables.
Such insights, which go beyond global importance rankings, are critical for identifying and interpreting process-level interactions that contribute to quality degradation. By systematically uncovering these compound effects, CLAIRE enables practitioners to move from abstract model interpretability toward actionable process knowledge. This reinforces the framework’s potential not only as a predictive tool but also as a diagnostic aid for complex industrial systems. The consistently dominant features and compound interaction patterns revealed by the latent exploration, such as feature pairs that jointly exert strong influence across multiple latent dimensions, represent precisely the type of diagnostic signals that can be leveraged in practice when domain knowledge is available. In real industrial deployments, where sensor metadata and process semantics are known, such interaction patterns can be interpreted by domain experts to associate latent behaviors with known fault modes or abnormal operating conditions. In this sense, the proposed interpretability layer is best viewed as a foundational bridge between representation learning and domain-level diagnosis, enabling the transition from statistical explanations to actionable process knowledge when combined with expert insight.

\section{Conclusions and Future Work}\label{section.conclusion}

This paper has presented CLAIRE, a hybrid end-to-end learning framework that integrates unsupervised deep representation learning with supervised classification for intelligent fault detection in industrial systems. The core of the architecture is an optimized encoding-decoding process that projects raw sensor data into a compact and structured latent space. By incorporating variance-aware reconstruction loss and latent regularization, CLAIRE captures subtle feature dependencies while mitigating the impact of noise and redundancy. The latent representations are passed to a kernel-based classifier. Empirical results demonstrate the superiority of the proposed model. In addition to robust feature extraction, CLAIRE addresses the interpretability challenge commonly associated with deep learning. A game-theoretic interpretability module is integrated to analyze latent space. This enables the identification of the most informative input features and their interactions, offering transparency in model behavior, which is an essential requirement in industrial environments. The insights gained from latent space not only support accurate fault prediction but also help reveal the root causes of quality degradation by attributing predictions to statistically relevant features. These properties make CLAIRE a practical and explainable AI-based solution for smart manufacturing. While the present study emphasizes an application-driven and empirically validated framework, a promising direction for future work lies in developing a formal theoretical characterization of latent variance regularization. Such theoretical developments would complement the empirical findings reported here and further strengthen the understanding of geometry-aware regularization mechanisms for industrial representation learning. Our future work also aims to extend this framework to handle multi-class and multi-label scenarios, as well as explore domain adaptation techniques \cite{YaoKang2023} for cross-factory generalization. The underlying idea of CLAIRE is also transferable to other domains, including healthcare, energy systems, and autonomous monitoring, where data complexity and explainability are equally critical.



\ifCLASSOPTIONcaptionsoff
  \newpage
\fi

\vspace{-20pt}
\begin{IEEEbiography}[{\includegraphics[width=1in,height=1.25in,clip,keepaspectratio]{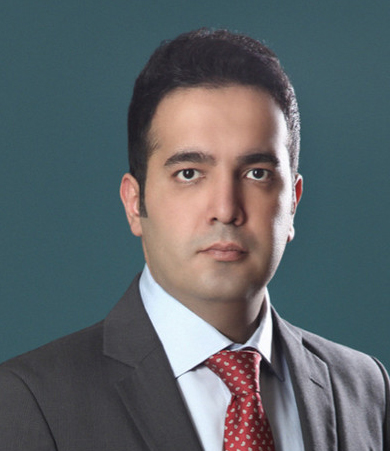}}]{Mohammadhossein Ghahramani}(Senior Member, IEEE) obtained his Ph.D. in Computer Technology and Application from Macau University of Science and Technology in 2018. He was a member of the Insight Centre for Data Analytics and a Research Fellow at University College Dublin, Ireland. Currently, he is an Assistant Professor of Data Science at Birmingham City University, UK. His research interests include smart systems, artificial intelligence, optimization, smart cities, and IoT. Dr Ghahramani has published numerous papers in reputable journals and has received several awards. He serves as a co-chair of the IEEE SMCS Technical Committee on AI-based Smart Manufacturing Systems and as an Associate Editor of IEEE Internet of Things Journal.\end{IEEEbiography}

\vspace{-15pt}
\begin{IEEEbiography}[{\includegraphics[width=1in,height=1.25in,clip,keepaspectratio]{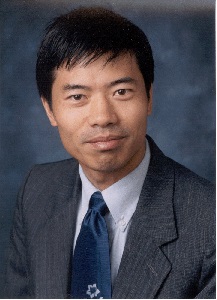}}]{MengChu Zhou}
(S'88-M'90-SM'93-F'03)(Fellow, IEEE) received his Ph. D. degree from Rensselaer Polytechnic Institute, Troy, NY in 1990 and then joined New Jersey Institute of Technology where he has been Distinguished Professor since 2013. His interests are in Petri nets, automation, robotics, big data, Internet of Things, cloud/edge computing, and AI.  He has 1400+ publications including 17 books, 900+ journal papers (700+ in IEEE transactions), 31 patents and 32 book-chapters. He is Fellow of IFAC, AAAS, CAA and NAI.
\end{IEEEbiography}


\end{document}